\documentclass{article}
\usepackage{graphicx}
\usepackage{amsmath}
\usepackage{xcolor}

\title{Forward-Cooperation-Backward (FCB) learning in a Multi-Encoding Uni-Decoding neural network architecture}
\author{Prasun Dutta\\
Machine Intelligence Unit\\
Indian Statistical Institute\\
Kolkata, India\\
{\tt\small amiprasun@gmail.com}
\and
Koustab Ghosh\\
Machine Intelligence Unit\\
Indian Statistical Institute\\
Kolkata, India\\
{\tt\small koustab2000@gmail.com}
\and
Rajat K. De\\
Machine Intelligence Unit\\
Indian Statistical Institute\\
Kolkata, India\\
{\tt\small rajat@isical.ac.in}
}
\date{}

\begin{document}
\maketitle

\begin{abstract}
The most popular technique to train a neural network is backpropagation. Recently, the Forward-Forward technique has also been introduced for certain learning tasks. However, in real life, human learning does not follow any of these techniques exclusively. The way a human learns is basically a combination of forward learning, backward propagation and cooperation. Humans start learning a new concept by themselves and try to refine their understanding hierarchically during which they might come across several doubts. The most common approach to doubt solving is a discussion with peers, which can be called cooperation. Cooperation/discussion/knowledge sharing among peers is one of the most important steps of learning that humans follow. However, there might still be a few doubts even after the discussion. Then the difference between the understanding of the concept and the original literature is identified and minimized over several revisions. Inspired by this, the paper introduces Forward-Cooperation-Backward (FCB) learning in a deep neural network framework mimicking the human nature of learning a new concept. A novel deep neural network architecture, called Multi Encoding Uni Decoding neural network model, has been designed which learns using the notion of FCB. A special lateral synaptic connection has also been introduced to realize cooperation. The models have been justified in terms of their performance in dimension reduction on four popular datasets. The ability to preserve the granular properties of data in low-rank embedding has been tested to justify the quality of dimension reduction. For downstream analyses, classification has also been performed. An experimental study on convergence analysis has been performed to establish the efficacy of the FCB learning strategy.

\end{abstract}

\section{Introduction}
\label{introduction_meud}
Neural networks have proven to be a key component of artificial intelligence. It has a wide area of applications like pattern recognition, image classification, dimensionality reduction, and natural language processing, among others. An essential aspect of a neural network is its learning technique which enables the network to learn patterns on its own, adapt to new data and provide efficient representations. Learning is a process of estimating weight parameters, based on the input and output of the network, to meet certain objectives.

The predominant method for training neural networks is backpropagation. The phrase ``backpropagation" refers to a technique for efficiently computing the gradient, not how the gradient is employed. The phrases ``reverse mode of automatic differentiation" or ``reverse accumulation" \cite{linnainmaa1970representation} are also used in place of ``backpropagation" \cite{goodfellow2016deep}. Backpropagation, an application of the chain rule of differential calculus to neural networks, is sometimes used informally to refer to the whole learning procedure. Backpropagation efficiently computes the gradient of a loss function with respect to the weights of the network at one layer at a time and proceeding backwards from the last layer. This way of proceeding backwards avoids repeating calculations of the intermediate terms in the chain rule, which can be obtained through dynamic programming \cite{kelley1960gradient, bryson1961gradient, goodfellow2016deep}.

The traditional method of learning by a neural network using forward and backward passes of backpropagation, storing the error derivatives layer-by-layer has been observed to be biologically implausible. Thus, Hinton developed The Forward Forward (FF) Algorithm \cite{hinton2022forward}, a novel greedy multi-layer learning procedure, influenced by Boltzmann machines \cite{hinton1986learning} and Noise Contrastive Estimation \cite{gutmann2010noise}. The FF algorithm for neural networks discards backpropagation across the network and learns using two layer-wise forward passes.

\paragraph{The Forward Forward Algorithm:}
It has been observed that the way a conventional neural network learns, storing the error derivatives layer by layer and using forward and backward passes of backpropagation, is not biologically plausible. Hinton thus developed a new technique, called The Forward Forward (FF) Algorithm \cite{hinton2022forward} for neural networks to learn with two forward passes applied in a layer-wise fashion, discarding backpropagation on the entire network. The FF algorithm deals with two types of data, namely, positive data and negative data. Positive data are real observed data, whereas negative data are generated by the model for better representation learning. The first forward pass is a positive pass on the positive samples and the second forward pass is a negative pass on the negative samples. The positive pass adjusts the weights to increase the value of the goodness term in each hidden layer, whereas the negative pass adjusts the weights to decrease the same. For every hidden layer, the goodness term is calculated as the squared sum of the output of each neuron of that layer. The FF algorithm aims to correctly distinguish between positive and negative data in every hidden layer. The probability of concluding that the data is positive is given by
\begin{equation}
 p(positive) = \sigma(\sum_{j}y^2_j-\theta)   
\end{equation}
where $p(positive)$ is the probability of determining if a sample is positive, $\theta$ is a threshold to regulate the value of the goodness and $y_j$ is the output of hidden node $j$. The layer-wise training/learning tries to maintain the goodness term well above the threshold value ($\theta$) for positive data and well below $\theta$ for the negative ones. After distinguishing the positive and negative values in the current hidden layer, the layer undergoes normalization to go forward for further processing by the next hidden layer. Layer normalization ensures that the next hidden layer is not influenced by the predictions of the current hidden layer. The final hidden layer is connected to the output layer for final prediction.

Researchers have tried to unify the advantages of backpropagation and the forward-forward algorithm to develop a more efficient learning technique. The FF algorithm has been generalized with predictive coding into a robust stochastic neural system by Ororbia and Mali \cite{ororbia2023predictive}. The Integrated Forward-Forward Algorithm (IntFF) \cite{tang2023integrated}, developed by Tang, leverages the design of multiple local loss functions to decompose the training target, eliminating the need to propagate gradients through the entire network structure. Instead, it employs shallow backpropagation, adjusting weights in just 1-3 layers of local hidden units. Lorberbom et al. have introduced inter-layer communication during training of an FF network which strengthens the effective communication between layers, thus leading to a better performance \cite{lorberbom2024layer}. The FF Algorithm has been unified with backpropagation through a Model Predictive Control (MPC) \cite{ren2024unifying} to incorporate feedback control over dynamic systems. Giampaolo et al. have used the idea of local backpropagation to create a hybrid network, thus avoiding the need for backward computations when needed \cite{giampaolo2023investigating}.  The superiority of FF algorithm in comparison to backpropagation in modeling the nature of learning in the cortex has been shown by Tosato et al. \cite{tosato2023emergent}. The extension of FF to CNN has been introduced in \cite{scodellaro2023training}. A recent variation of FF which does not require the generation of additional negative samples during training has been developed by Zhao et al. \cite{zhao2023cascaded}.

In this paper, we come up with a novel learning technique called Forward-Cooperation-Backward (FCB) learning. The FCB learning is philosophically based on the nature of learning humans follow while learning a new concept as a combination of forward-forward learning, cooperation and backpropagation.
To incorporate FCB learning, the novel Multi Encoding Uni Decoding (MEUD) neural network framework has been developed. The gradual development of MEUD, MEUD-FF, MEUD-Coop to MEUD-FF-Coop helps to completely realize the FCB learning mechanism.

The efficacy of the MEUD-FF-Coop framework has been demonstrated with a comparison against MEUD, MEUD-FF, MEUD-Coop and the standard deep Autoencoder. The dimensionally reduced data obtained from the aforementioned networks have been tested thoroughly. The ability to preserve the granular relationship between data in original and projected spaces has been evaluated through the trustworthiness score. The classification performance on the dimensionally reduced data has been judiciously evaluated using different well-known classification algorithms on a number of popular classification metrics. The convergence analysis has also been performed through an experimental study.

The remaining sections of the paper are organised as follows. The motivation, design and learning of the proposed deep learning models MEUD, MEUD-FF, MEUD-Coop and MEUD-FF-Coop have been discussed in Section~\ref{model_meud}. Section~\ref{experiments_meud} summarizes the datasets considered, it further demonstrates the technicalities of the experimental setup, experimental procedure, and presents the results with their corresponding analysis. The paper is concluded in Section~\ref{conclusion_meud}.  

\section{MEUD, MEUD-FF, MEUD-Coop and MEUD-FF-Coop}
\label{model_meud}
This section describes the philosophy behind the Forward-Cooperation-Backward (FCB) way of learning that has been inspired by the way human learns. The FCB way of learning has gradually been realized through the novel MEUD-FF-Coop architecture. This is followed by the detailed delineation of the MEUD, MEUD-FF, MEUD-Coop and MEUD-FF-Coop architectures. Additionally, this section discusses the learning strategies employed by each of these models.

\subsection{Motivation}
\label{motivation_meud}
When a new subject is introduced to the students of a class, the students go through several steps in order to master the same. The students start progressing chapter-wise to grasp the concepts. However, a student may not understand the concept of a chapter holistically at once. First, the student tries to learn the most important concepts of that chapter. The student may need to refine the learned concept several times. Thus, the student forwards hierarchically. After some steps, the student might understand some of the concepts and have a few doubts. Now, the student tries to discuss them with their friends. This phase of learning can be called cooperation. The discussion helps the student to clear most of the doubts, but might not all. For the remaining doubts, the student has to identify the difference between their learning and the actual content. By identifying the differences the student needs to refresh the previous knowledge and redo the above steps of learning. This process is nothing but backpropagation. Inspired by this human centric learning procedure, we have developed a neural network based learning strategy, called Forward-Cooperation-Backward (FCB) way of learning.

In order to realize the FCB learning strategy, a novel neural network architecture, called Multi Encoding Uni Decoding (MEUD) neural network, has been developed. The MEUD architecture starts with all the features of the input and proceeds forward layer-wise, processing information hierarchically. At every step of this process, the accumulated information gets refined. This part of the concept has been accomplished using the FF algorithm (MEUD-FF neural network). After a certain number of steps, the bottleneck occurs and the forward path gets choked. Here for further processing, the cooperation layer is introduced. In this cooperation layer, any particular node is connected only with its neighbours (MEUD-Coop neural network). This type of connection resembles cooperation among friends. We name this layer as the latent layer. After discussion among peers, to identify the remaining problems, the latent layer nodes directly refer back to the original feature space. At this time, the network needs to backpropagate. The network incorporating all these three concepts has been named MEUD-FF-Coop neural network.

Starting with the input layer of nodes, several hidden layers are there in the architecture designed for hierarchical learning. This hierarchical forward learning leads to the bottleneck layer of the network, where the cooperation takes place with the latent layer nodes. The latent representation of the input is generated at this point as the output of this latent layer. We can think that the processing of data from input to this layer is the encoding of information from its original feature space to its latent space. At this point, the comparison with the original feature space needs to take place, and for this, the latent space representation is projected back to the original space. Thus, in this latent to output layer portion, the architecture tries to decode the learned latent space representation back to the original space. Hence, there are multiple encoding layers and a single decoding layer in the architecture. Thus, the model has been named Multi Encoding Uni Decoding (MEUD) neural network.

In the next sections, we first introduce the MEUD architecture (Section~\ref{architecture_meud}) in detail. Going further, the process of incorporating the FF algorithm in the MEUD architecture has been delineated (Section~\ref{architecture_meud_FF}). Section~\ref{architecture_meud_Coop} describes the details of implementing cooperation in the MEUD network. Finally, the overall MEUD-FF-Coop neural network has been presented in Section~\ref{architecture_meud_FF_Coop}. The learning procedure of all these four networks along with the objective function has been described in Section~\ref{learning_meud}.

\subsection{Multi Encoding Uni Decoding (MEUD) neural network}
\label{architecture_meud}
\begin{figure}[t]
\centerline{
\includegraphics[width=\linewidth, scale=1.0]{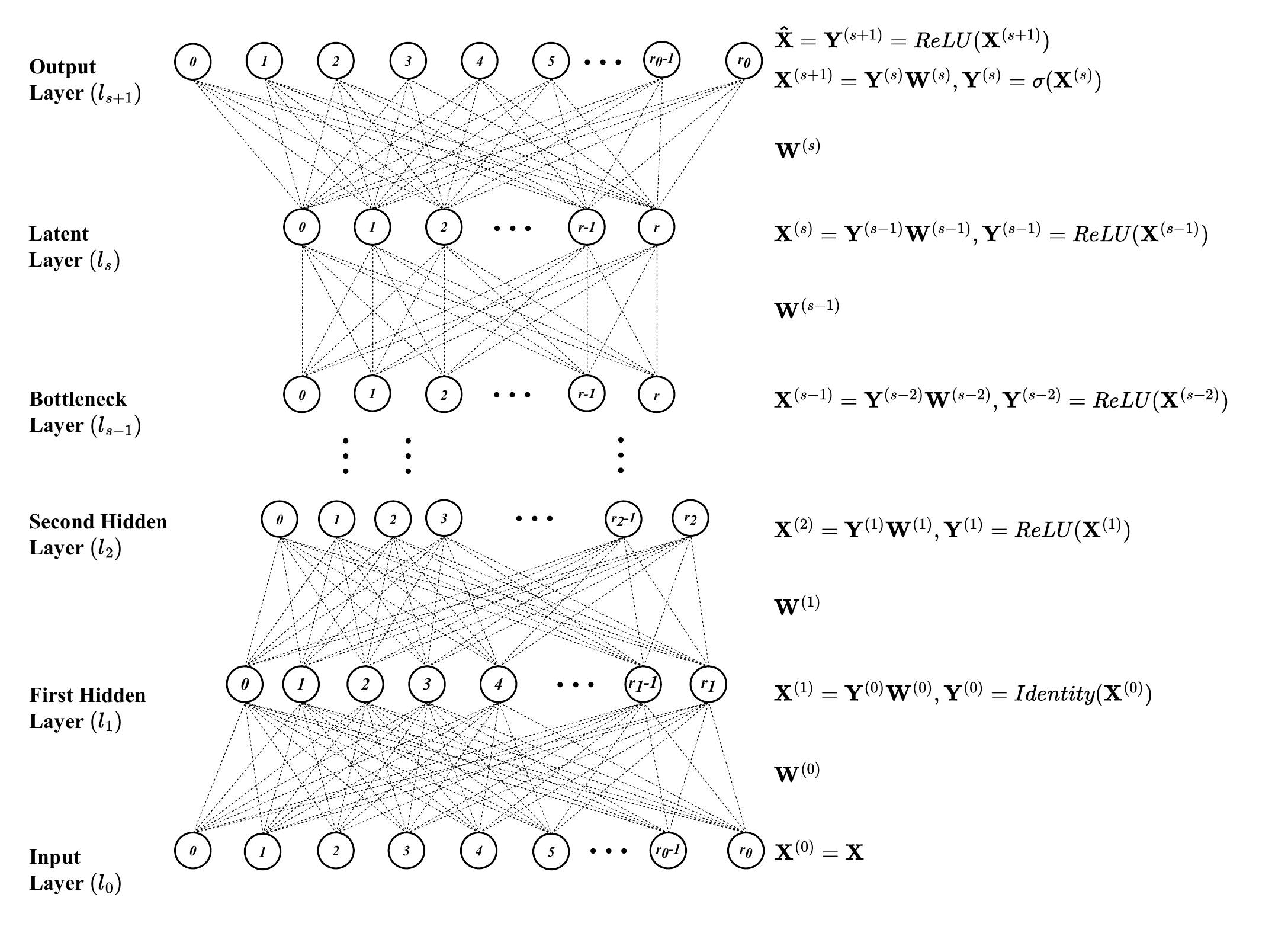}}
\caption{The architecture of MEUD neural network model}
\label{meud_architecture}
\end{figure}

The MEUD architecture attempts to obtain a dimensionally reduced representation of the original input data $\mathbf{X}=[x_{ij}]_{m \times n}$, where $x_{ij}$ represents the $j^{th}$ feature value of the $i^{th}$ sample. As depicted in Figure~\ref{meud_architecture}, the MEUD network comprises an input layer, $s$ hidden layers and an output layer, i.e., a total of $(s+2)$ layers. The input layer $(l_0)$ has $r_{0}=n$ nodes and receives the input ($\mathbf{X}$) to the architecture. The first hidden layer $(l_1)$ has $r_{1}<r_{0}$ nodes and the second hidden layer $(l_2)$ has $r_{2}<r_{1}$ nodes. Proceeding in this fashion, the $(s-1)^{th}$ hidden layer ($l_{s-1}$) having $r_{s-1}<r_{s-2}$ nodes has been designed to be the bottleneck layer of the framework. The bottleneck layer ($l_{s-1}$) has been projected to the final hidden layer, which has been designated as the latent layer ($l_s$) of the architecture. The desired low-rank approximation of the input data is obtained as the output of this layer. The latent layer and the bottleneck layer have been designed to have the same number of nodes,i.e., $r_{s}=r_{s-1}=r$, where $r$ is the desired target dimension. The latent layer is further projected to the output layer $(l_{s+1})$ having $r_{s+1}=n$ nodes. The aim of this output layer is to reconstruct the input to the model.

Based on the various requirements of MEUD, three different types of activation functions, viz., Identity, Sigmoid and ReLU have been used. The job of the input layer is to pass on the input without any processing to the next layer of the architecture. For this, the Identity function has been used. The sigmoid function $(\sigma)$ maps any input to the range $(0, 1)$, and thus has been used in the latent layer of the architecture, producing the low-rank approximation of the input matrix. For all other layers, the ReLU activation function has been used.

The input to the $k^{th}$ layer of MEUD is represented as $\mathbf{X}^{(k)}$ for $k=0,1,2,...,(s+1)$. The output of the $k^{th}$ layer is represented as $\mathbf{Y}^{(k)}$ for $k=0,1,2,...,(s+1)$. For the output layer, i.e., when $k=s+1$, $\mathbf{Y}^{(s+1)}=\widehat{\mathbf{X}}$. We compute $\mathbf{Y}^{(k)}$ as
\begin{equation}
\mathbf{Y}^{(k)} = 
\begin{cases}
        Identity(\mathbf{X}^{(0)}), & \text{for}\ k=0\\
        \sigma(\mathbf{X}^{(s)}), & \text{for}\ k=s\\
        ReLU(\mathbf{X}^{(k)}), & \text{otherwise}
\end{cases}
\label{y_k_meud}
\end{equation}
$\mathbf{W}^{(k)}$, for $k=0,1,2,...,s$, represents the weight matrix between two consecutive layers $l_k$ and $l_{(k+1)}$. Thus, $\mathbf{X}^{(k)}$ for $k=0,1,2,...,(s+1)$ can be computed as
\begin{equation}
\mathbf{X}^{(k)} = 
\begin{cases}
        \mathbf{X}, & \text{for}\ k=0\\
        \mathbf{Y}^{(k-1)}\mathbf{W}^{(k-1)}, & \text{otherwise}
\end{cases}
\label{x_k_meud}
\end{equation}
The MEUD architecture has been designed to initialize all the weight matrices using samples drawn from a standard random normal distribution with mean $0$ and standard deviation $0.1$.

\subsection{MEUD-FF neural network}
\label{architecture_meud_FF}
The architecture of the MEUD-FF neural network is the same as that of the MEUD neural network, having a total of $(s+2)$ layers (Figure~\ref{meud_architecture}). The MEUD and MEUD-FF architecture differs in the weight initialization strategy of the first $s$ layers of the network. In the MEUD architecture, the initial value of the weights is drawn from a standard random normal distribution. In the case of MEUD-FF the first $s$ layers i.e., layers $l_0$ to $l_{s-1}$ have been designed to be initialized by the trained weights of the $s$ shallow FF networks. The weights of the remaining two layers, i.e., layers $l_s$ and $l_{s+1}$ have been initialized using the same strategy as that of MEUD.

A shallow FF model comprises two layers, namely, the input and a hidden layer. The hidden layer has been designed to serve as the output layer of the shallow FF model. The input to the MEUD-FF ($\mathbf{X}^{(0)}$), i.e., the output of the input layer $l_0$ denoted by $\mathbf{Y}^{(0)}$ is passed to the first shallow FF model for training. The trained weights of the first shallow FF model are used to initialize the weight matrix $\mathbf{W}^{(0)}$, connecting $l_0$ and $l_1$ layers of the MEUD-FF network. The output of the first hidden layer ($\mathbf{Y}^{(1)}$) is computed using equation~(\ref{y_k_meud}). Now, $\mathbf{Y}^{(1)}$ serves as the input to the second shallow FF model, which is further trained and the trained weights are used to initialize $\mathbf{W}^{(1)}$, connecting layers $l_1$ and $l_2$. The same fashion of weight initialization is followed for the remaining weight matrices up to the $s^{th}$ hidden layer of the MEUD-FF network.

\subsection{MEUD-Coop neural network}
\label{architecture_meud_Coop}
\begin{figure}[t]
\centerline{
\includegraphics[width=\linewidth, scale=1.0]{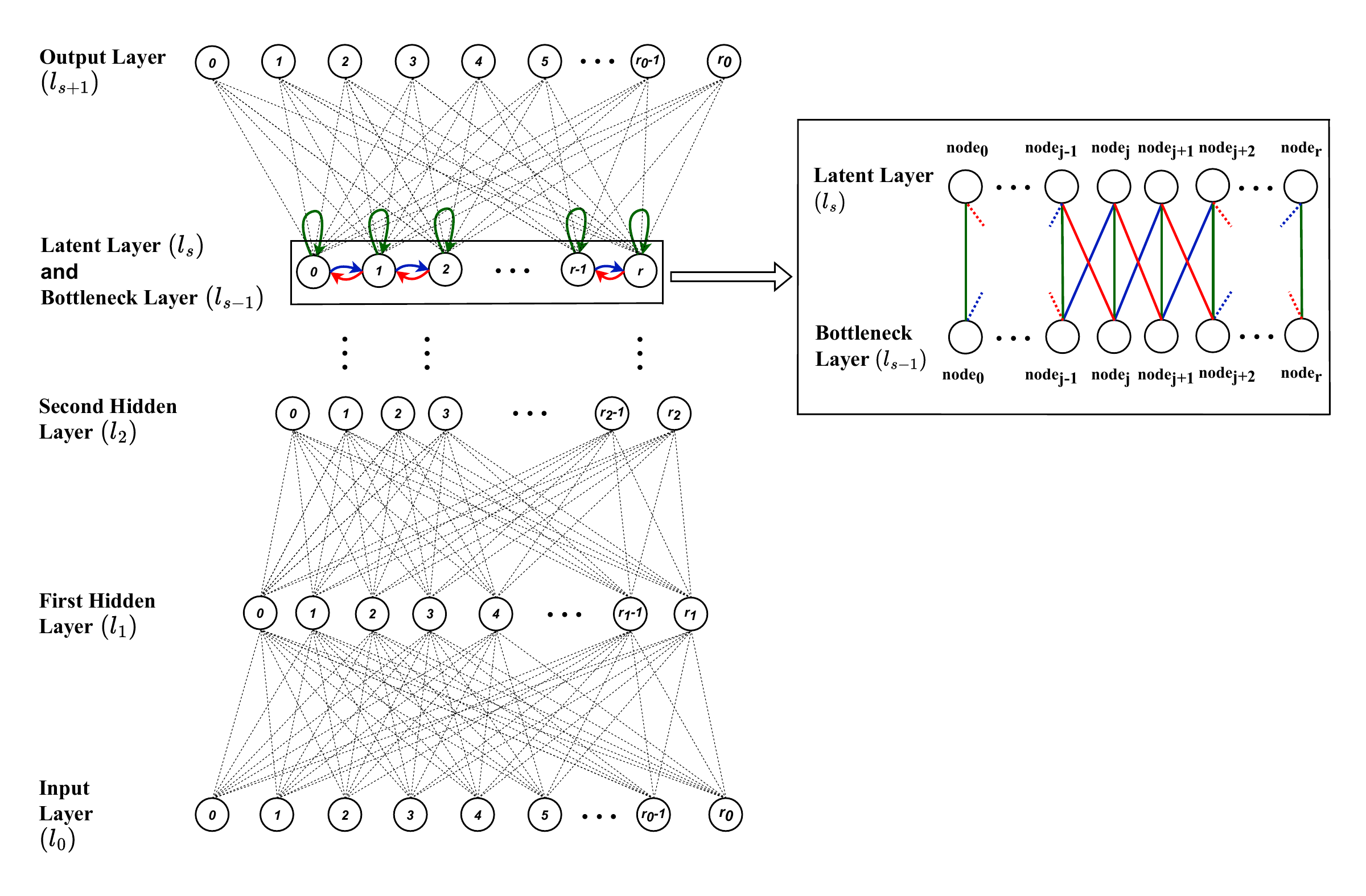}}
\caption{The architecture of MEUD-Coop neural network model}
\label{meud_coop_architecture}
\end{figure}

The MEUD-Coop follows the same architectural properties as MEUD. The differentiating factor lies in the conjunction of the bottleneck and the latent layer. Both the layers have the same number of nodes $r$. Philosophically, these two layers act as a single layer of nodes with specific synaptic connections between them. The bottleneck/latent layer undergoes cooperation among the neighbouring nodes as depicted in Figure~\ref{meud_coop_architecture}. The $j^{th}$ node ($node_j$) of the layer is connected to itself, $(j-1)^{th}$ node ($node_{j-1}$) and $(j+1)^{th}$ node ($node_{j+1}$) of the same layer. The self-connection of a node is denoted by the colour green in Figure~\ref{meud_coop_architecture}. The blue and red colours denote the left and right synaptic connections of a node respectively. For a better visualization of these special synaptic connections between the nodes of the bottleneck/latent layer, an enhanced depiction of this portion of the MEUD-Coop framework has been presented in Figure~\ref{meud_coop_architecture}. These special connections help to realize the novel concept of cooperation introduced in the MEUD-Coop neural network architecture. The weight initialization strategy of MEUD-Coop architecture is the same as that of the MEUD architecture.

\subsection{MEUD-FF-Coop neural network}
\label{architecture_meud_FF_Coop}
To realize the Forward-Cooperation-Backward learning as described in Section~\ref{motivation_meud}, the MEUD-FF-Coop neural network architecture has been designed. MEUD-FF-Coop is a combination of the MEUD-FF and MEUD-Coop models. The framework of the MEUD-FF-Coop architecture is the same as that of MEUD-Coop (Figure~\ref{meud_coop_architecture}) incorporating special neighbouring connections among the bottleneck/latent layer nodes. The weight initialization strategy of the MEUD-FF-Coop neural network is the same as that of the MEUD-FF architecture. Among the total $(s+2)$ layers of MEUD-FF-Coop architecture, the weight matrices connecting the first $s$ layers are initialized from the learned weights of the $s$ shallow FF models and the remaining weight matrices connecting the bottleneck, latent and output layers are initialized using the weights taken from a standard random normal distribution. The output layer tries to reconstruct the input to the model, and thus paves the way for fine-tuning the weights of the entire network through backpropagation. This completes the realization of the envisioned FCB learning strategy of a neural network architecture.

\subsection{Learning}
\label{learning_meud}
The MEUD, MEUD-FF, MEUD-Coop, MEUD-FF-Coop aims to obtain a reconstruction ($\widehat{\mathbf{X}}$) of the original input $(\mathbf{X})$. For this purpose, we use the mean squared error loss between the original input data and the reconstructed output data. We aim to minimize $||\mathbf{X}-\widehat{\mathbf{X}}||_F$. The cost function $\mathbf{C}$ is defined as,
\begin{equation}
    \mathbf{C} = \frac{1}{2mn}\sum_{i=1}^{m}\sum_{j=1}^{n}(x_{ij} - \widehat{x}_{ij})^{2}
\end{equation}
MEUD, MEUD-FF, MEUD-Coop, MEUD-FF-Coop have been trained using the ADAM optimization technique \cite{kingma2014adam}.

\section{Experiments, Results and Analysis}
\label{experiments_meud}
The efficacy of the proposed FCB learning strategy mimicking the human way of learning new concepts accoutered on the novel MEUD-FF-Coop has been established demonstrating its supremacy in dimensionality reduction over the others. The step-by-step development of the MEUD-FF-Coop neural network model starting from the standard deep autoencoder to the novel MEUD architecture, from there to MEUD-FF and MEUD-Coop framework, and finally attaining the MEUD-FF-Coop model has been tested in terms of their competency in dimension reduction.

The effectiveness of dimensionality reduction has been examined through three primary approaches. First, we have compared the ability of various models to preserve the local structure of the data after reduction. Second, we have evaluated the quality of dimensionality reduction based on its impact on downstream analyses, specifically in classification tasks. Third, we have conducted an experimental convergence analysis of all five models. In our study, we have considered four well-known datasets to assess these dimensionality reduction techniques.

Section \ref{data_meud} mentions the data sources used. The experimental setup, weight initialization technique and the steps involved in data preparation have been provided in Section \ref{exp_setup_meud}. The magnitude of local structure preservation has been checked and the methodology used has been described in Section \ref{exp_procedure_meud}. The experimental procedure for determining the quality of dimensionality reduction in terms of classification has been described in Section \ref{exp_procedure_meud}. Section \ref{analysis_meud} is a comparative performance of all five models following the same experimental procedure described in \ref{exp_procedure_meud}. To conclude the convergence analysis, Section \ref{convergence_meud} depicts a graphical representation of the convergence of the objective function for all the neural network models compared here. 

\subsection{Data Sources}
\label{data_meud}
Four popular datasets namely, MNIST \cite{deng2012mnist}, Fashion MNIST \cite{xiao2017fashion}, CIFAR-10 \cite{krizhevsky2009learning}, and Extended MNIST \cite{cohen2017emnist} have been used to demonstrate the effectiveness of the FCB learning strategy. This strategy is evaluated in terms of dimensionality reduction using various models, including MEUD, MEUD-FF, MEUD-Coop, and MEUD-FF-Coop neural networks, compared against a standard deep autoencoder.

\subsubsection{MNIST Dataset}
The MNIST (Modified National Institute of Standards and Technology) dataset \cite{deng2012mnist} is a large collection of handwritten digits from $0-9$. Comprising of $60,000$ training and $10,000$ testing examples, each image is a grayscale image with $28\times28$ pixels. Each image is associated with a label from $0-9$ indicating the digit it represents.

\subsubsection{Fashion MNIST Dataset}
The Fashion MNIST (FMNIST) dataset \cite{xiao2017fashion} was released by Zalando, an online fashion retailer, with the goal of promoting more meaningful advancements in image recognition beyond the simple digits of MNIST. FMNIST contains $70,000$ grayscale images of various fashion items from $10$ different classes (e.g., T-shirts, bags, sneakers, etc.). FMNIST is split into $60,000$ training and $10,000$ testing examples each having $28\times28$ pixels.

\subsubsection{CIFAR-10 Dataset}
The CIFAR-10 dataset \cite{krizhevsky2009learning} was developed by the Canadian Institute for Advanced Research (CIFAR) to measure the effectiveness of image recognition algorithms. CIFAR-10 is challenging due to its small image size and a wide variety of objects and animals as classes. CIFAR-10 consists of $60,000$ color images categorized into $10$ distinct classes. The dataset is split into $50,000$ training images and $10,000$ testing images. Having $32\times32$ pixels, each image is composed of $3$ colour channels (RGB).

\subsubsection{Extended MNIST Dataset}
The Extended MNIST (EMNIST) \cite{cohen2017emnist} dataset is an extension of the original MNIST dataset, designed to broaden the scope of handwritten character recognition tasks. It includes both letters and digits as grayscale images. EMNIST contains 814,255 images of handwritten digits and characters across several subsets. It is split into training and testing sets of varying sizes, depending on the subset used. EMNIST is organized into six subsets, each with different character classes and counts, among which we have used the subset Letters comprising $26$ classes (A-Z, uppercase only), having $103,600$ images out of which  $88,800$ examples have been used for training and $14,800$ have been used for testing. Each grayscale image has $28\times28$ pixels. 

\subsection{Experimental setup}
\label{exp_setup_meud}
The MEUD, MEUD-FF, MEUD-Coop, and MEUD-FF-Coop have been implemented on a Python version of $3.10.15$ and a PyTorch version of $2.4.1+cu118$. The scikit-learn, Matplotlib and seaborn libraries have been used for various downstream analyses and plots.

The input data matrices have been preprocessed before being fed to the neural networks. The first $p$ pixels of each image have been replaced with a one-hot encoding of the labels, where $p$ denotes the number of classes of the dataset. To create the positive or real samples, the one-hot encoding is performed with the actual label associated with the image. The negative samples have been created using a one-hot encoding of any label but the original one associated with the image. For example, if we consider an image from the MNIST dataset representing $3$, the positive label of the image will be $3$, whereas the negative label will be drawn randomly from the set $\{0-9\}/\{3\}$. The images are then flattened and appended row-wise to create the input matrix $\mathbf{X}$. Considering a total of $m$ samples in $\mathbf{X},$ the first $m/2$ samples represent positive samples and the remaining are negative samples. All the dimensionality reduction techniques used here have been trained on the $m$ samples. However, while performing the classification tasks, each classifier has been trained on the positive $m/2$ dimensionally reduced samples. 

Each of the four datasets has been reduced to $20$ different $r$ dimensions by all five dimension reduction techniques. The value of $r$ ranges from $25$ to $500$ with two consecutive $r$ values differing by $25$. The average of these $20$ performance scores corresponding to twenty $r$ values has been presented in Section~\ref{analysis_meud}.

\subsection{Experimental Procedure}
\label{exp_procedure_meud}
The performance of MEUD-FF-Coop with respect to dimensionality reduction, in contrast to the progressive development from standard deep Autoencoder (AE) to MEUD, MEUD-FF and MEUD-Coop has been measured in two ways. First, the quality of local structure retention has been measured using the trustworthiness metric. Second, downstream analyses have further been performed for classification. Let $T$ be a set containing the five aforementioned dimensionality reduction techniques. A single dimensionality reduction technique is denoted as $\dot{T}$. A dataset $\mathbf{X}$, having $m$ samples and $n$ features per sample, is reduced to a dimension $r$ ($r<n$). The dataset $\mathbf{X}$ is dimensionally reduced using all $\dot{T}$ in $T$ for a certain value of $r$. Thus for a particular value of $r$, we obtain five dimensionally transformed datasets denoted by $\mathbf{X}_r(\dot{T})$. The efficiency of MEUD-FF-Coop will now be illustrated on these transformed datasets.

\begin{figure}[t]
\centerline{
\includegraphics[width=\linewidth, scale=1.0]{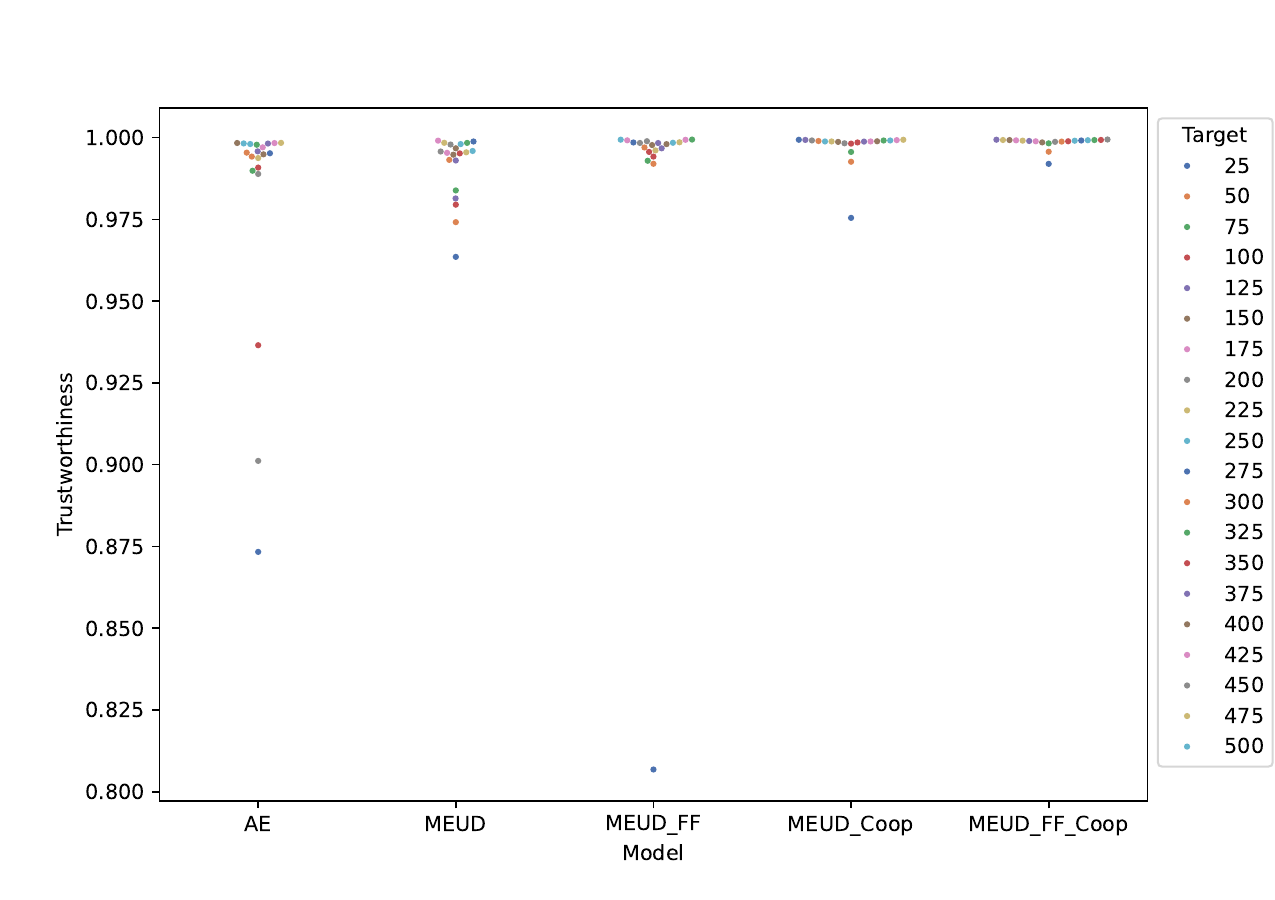}}
\caption{Trustworthiness scores of five deep neural network techniques on MNIST dataset for $20$ different $r$ values.}
\label{trustworthiness_MNIST}
\end{figure}

\begin{figure}[t]
\centerline{
\includegraphics[width=\linewidth, scale=1.0]{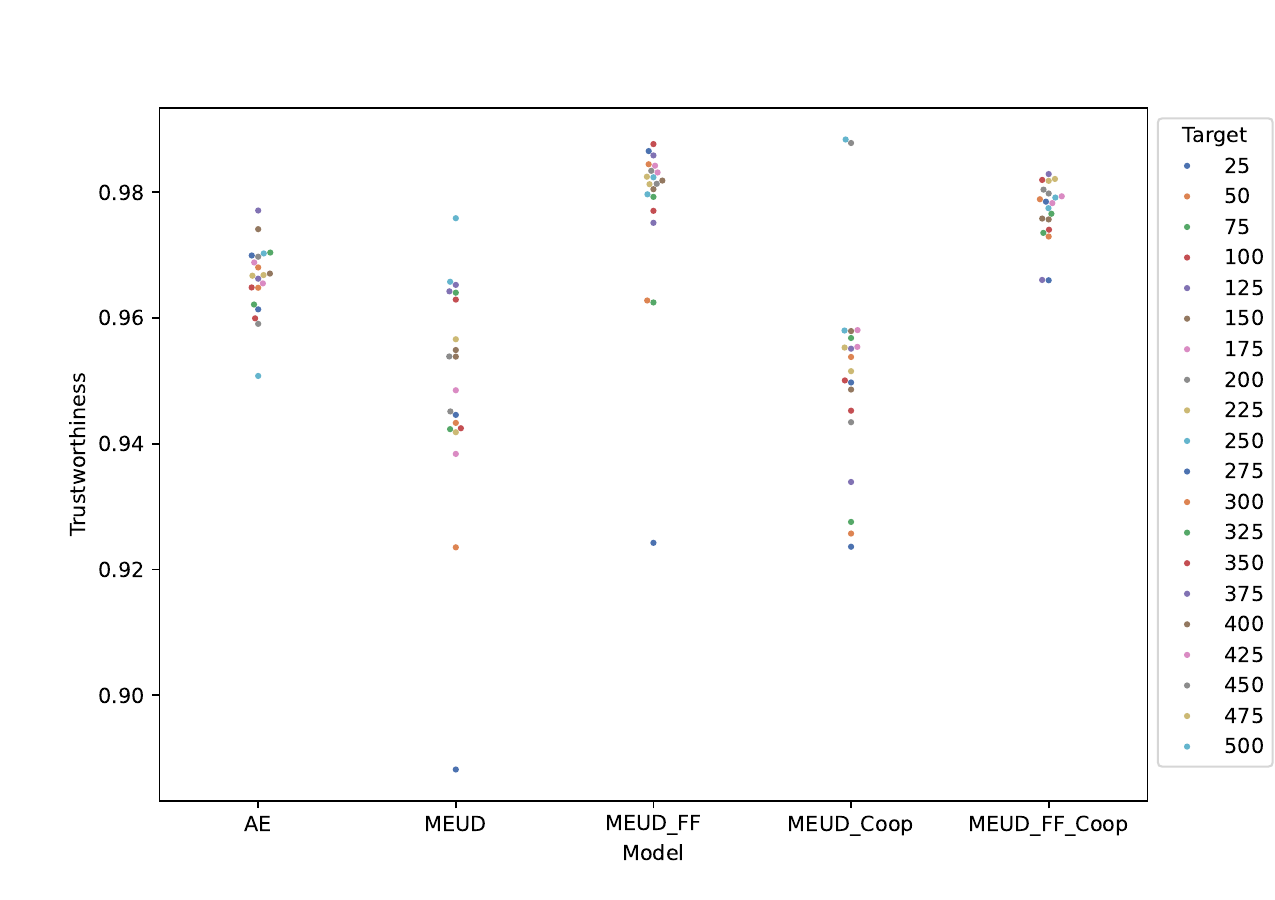}}
\caption{Trustworthiness scores of five deep neural network techniques on FMNIST dataset for $20$ different $r$ values.}
\label{trustworthiness_FMNIST}
\end{figure}

\begin{figure}[t]
\centerline{
\includegraphics[width=\linewidth, scale=1.0]{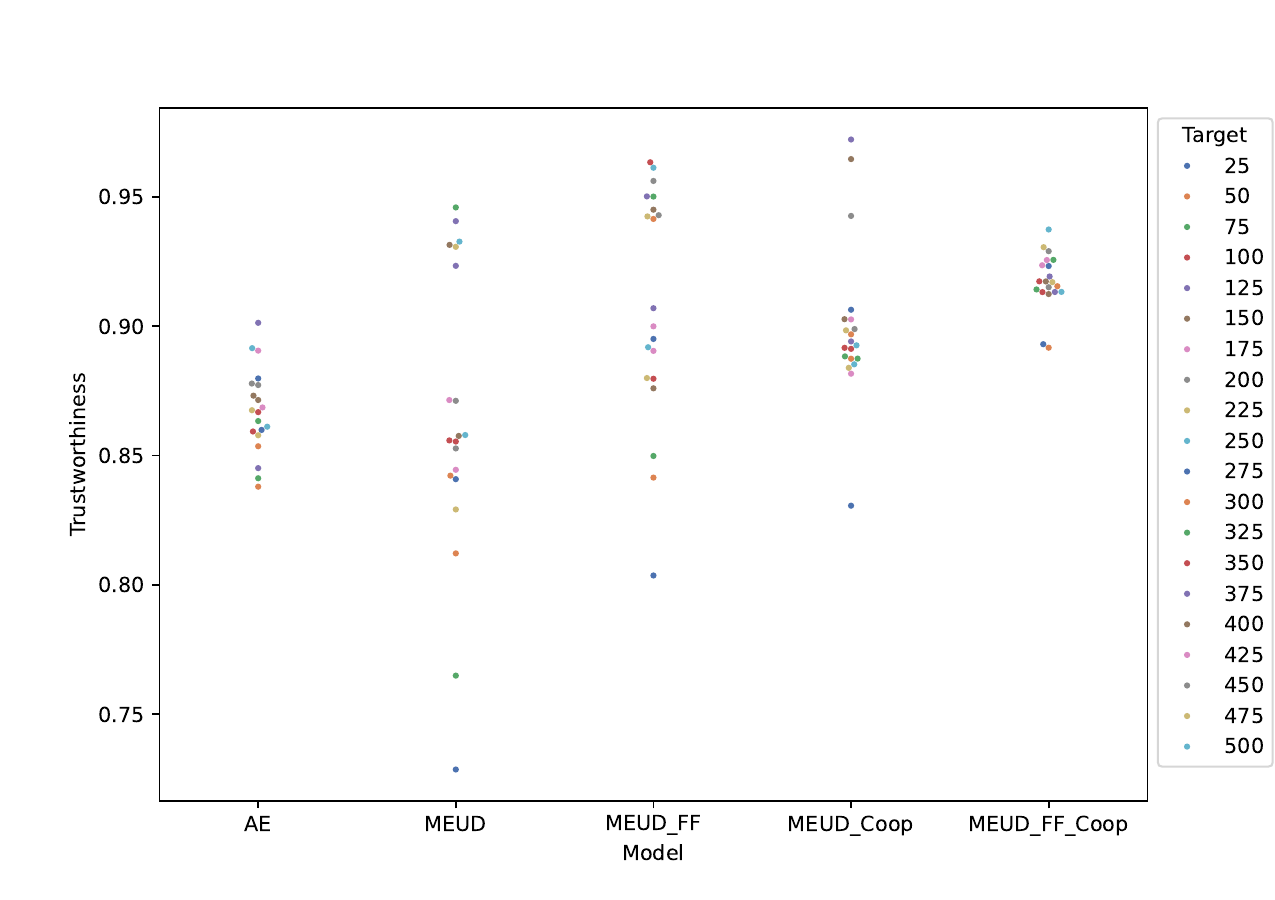}}
\caption{Trustworthiness scores of five deep neural network techniques on CIFAR-10 dataset for $20$ different $r$ values.}
\label{trustworthiness_CIFAR10}
\end{figure}

\begin{figure}[t]
\centerline{
\includegraphics[width=\linewidth, scale=1.0]{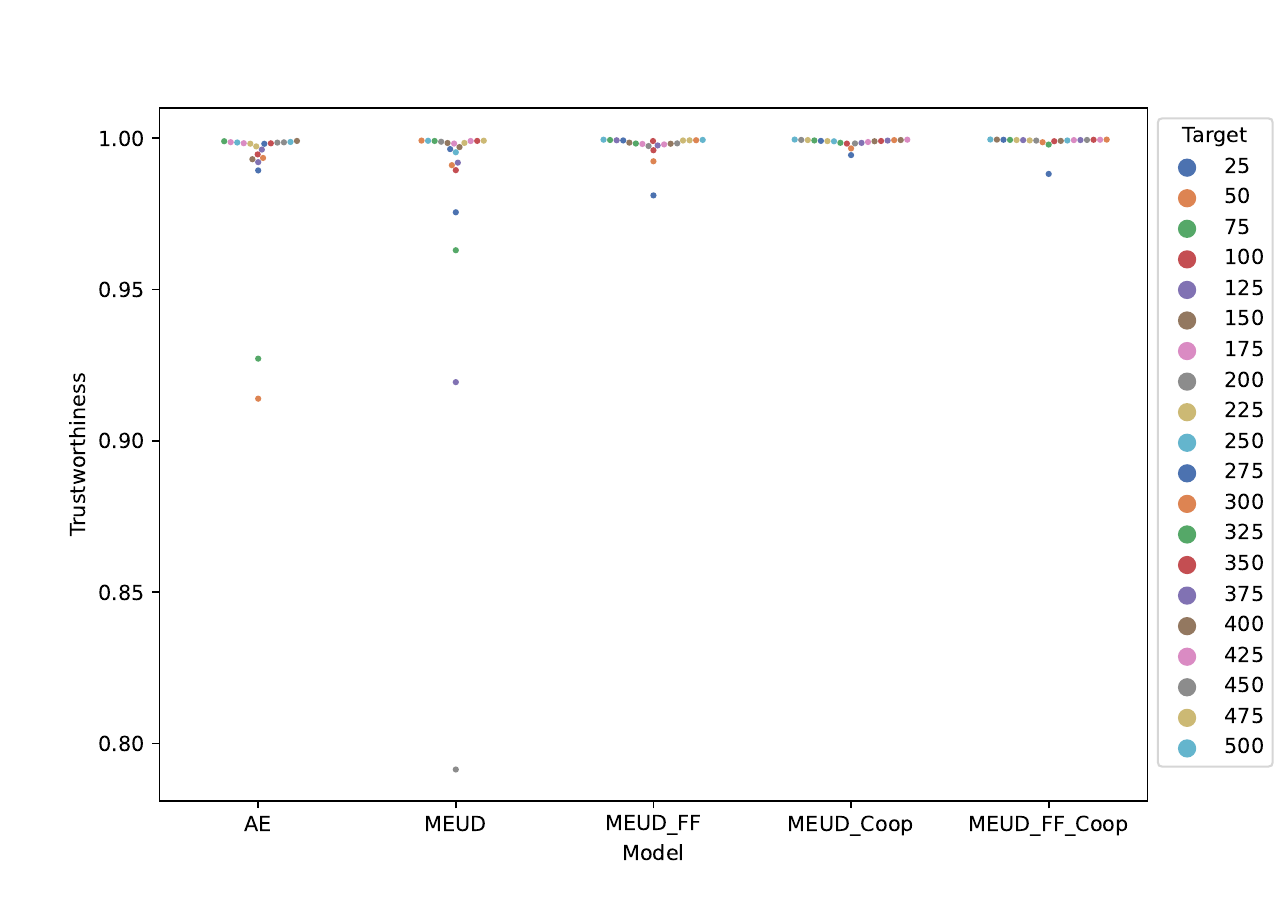}}
\caption{Trustworthiness scores of five deep neural network techniques on EMNIST dataset for $20$ different $r$ values.}
\label{trustworthiness_EMNIST}
\end{figure}

\begin{figure}[t]
\centerline{
\includegraphics[width=\linewidth, scale=1.0]{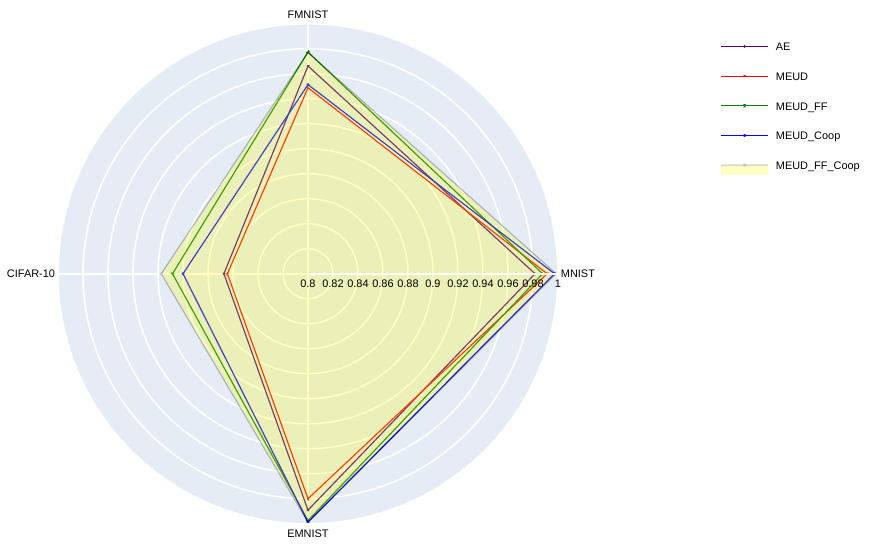}}
\caption{Mean trustworthiness scores of five deep neural network techniques including MEUD-FF-Coop.}
\label{trustworthiness}
\end{figure}

The downstream analyses are based on the classification of each reduced dataset $\mathbf{X}_r(\dot{T})$. Five well-known classification methods namely, K-Nearest Neighbor (KNN), Gaussian Naive Bayes (GNB), Logistic Regression (LRegression), XGBoost (XGB) and Multilayer Perceptron (MLP) have been used for classification. To evaluate the quality of classification, six classification performance metrics namely, Accuracy, Precision, Recall, F1 score, Jaccard index and Area Under the Receiver Operating Characteristic Curve (ROC\_AUC) have been considered. We get a classification performance score for each $\mathbf{X}_r(\dot{T})$ by performing classification using a classifier and validating the outcome using a classification performance metric.

\begin{figure}[t]
\centerline{
\includegraphics[width=.93\linewidth, scale=1.0]{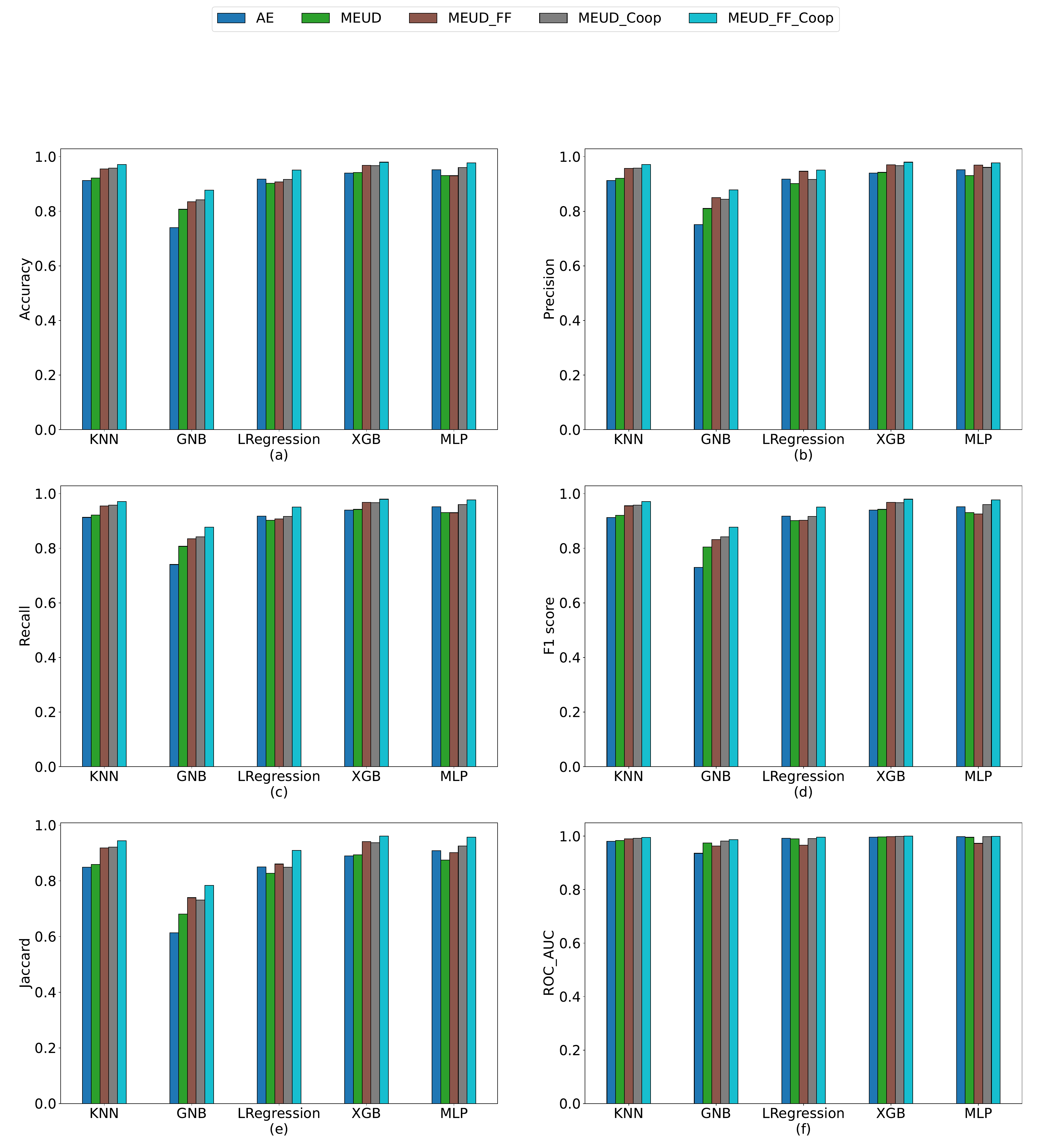}}
\caption{Mean performance scores of the classification algorithms on the dimensionally reduced dataset instances of the MNIST dataset by MEUD-FF-Coop and four other deep learning techniques.}
\label{classify_average_MNIST}
\end{figure}

\begin{figure}[t]
\centerline{
\includegraphics[width=.93\linewidth, scale=1.0]{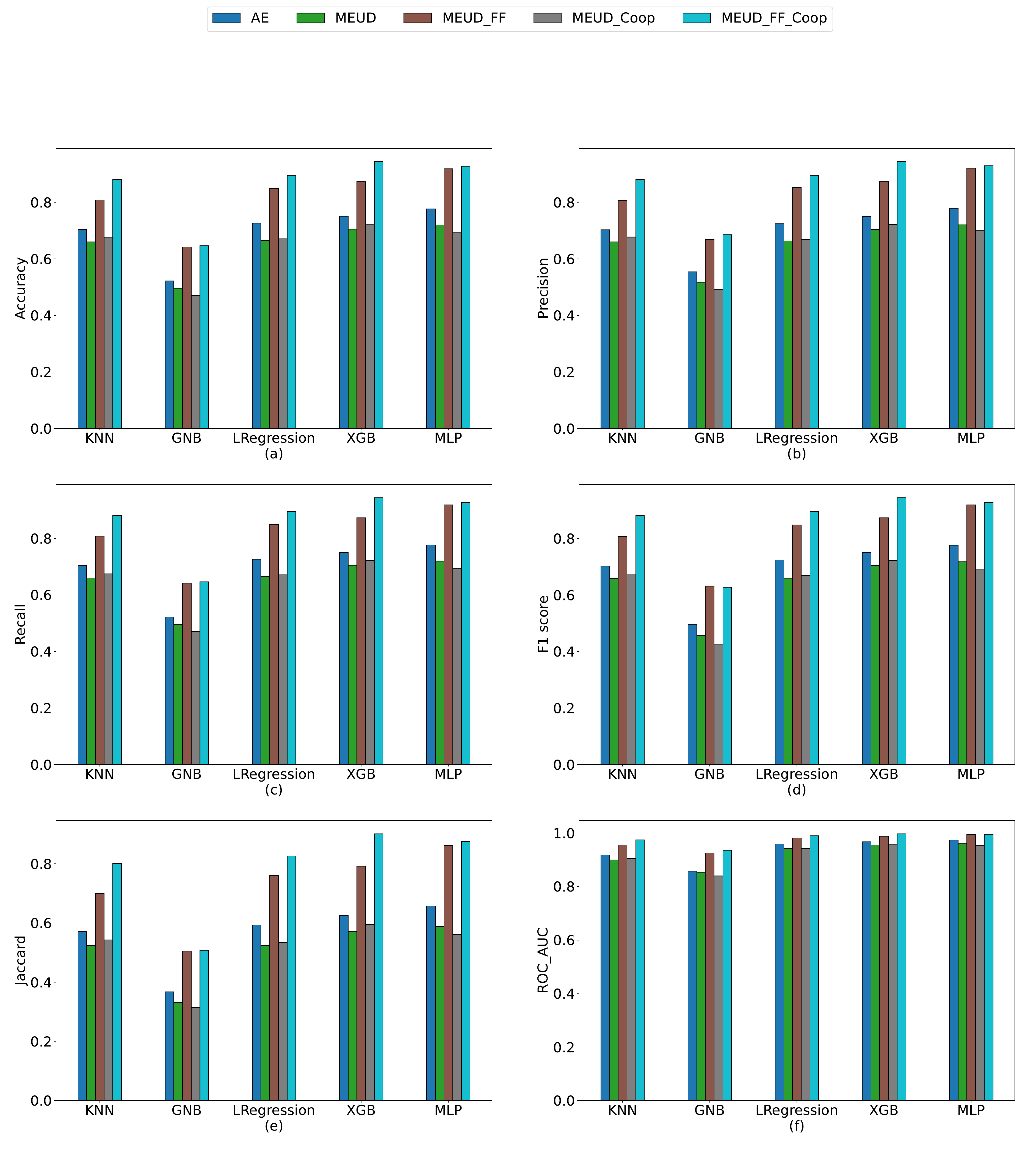}}
\caption{Mean performance scores of the classification algorithms on the dimensionally reduced dataset instances of the FMNIST dataset by MEUD-FF-Coop and four other deep learning techniques.}
\label{classify_average_FMNIST}
\end{figure}

\begin{figure}[t]
\centerline{
\includegraphics[width=.93\linewidth, scale=1.0]{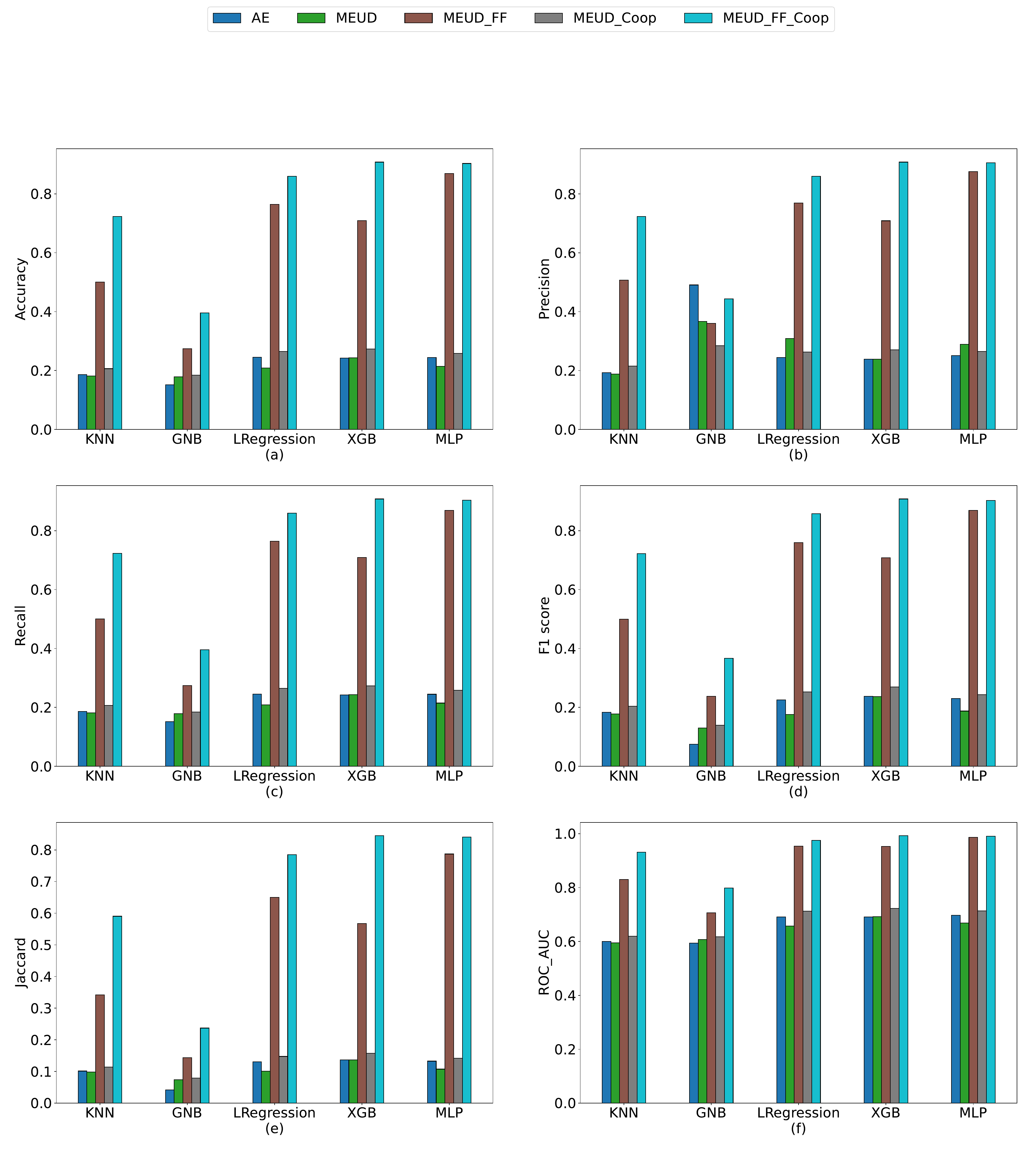}}
\caption{Mean performance scores of the classification algorithms on the dimensionally reduced dataset instances of the CIFAR-10 dataset by MEUD-FF-Coop and four other deep learning techniques.}
\label{classify_average_CIFAR10}
\end{figure}

\begin{figure}[t]
\centerline{
\includegraphics[width=.93\linewidth, scale=1.0]{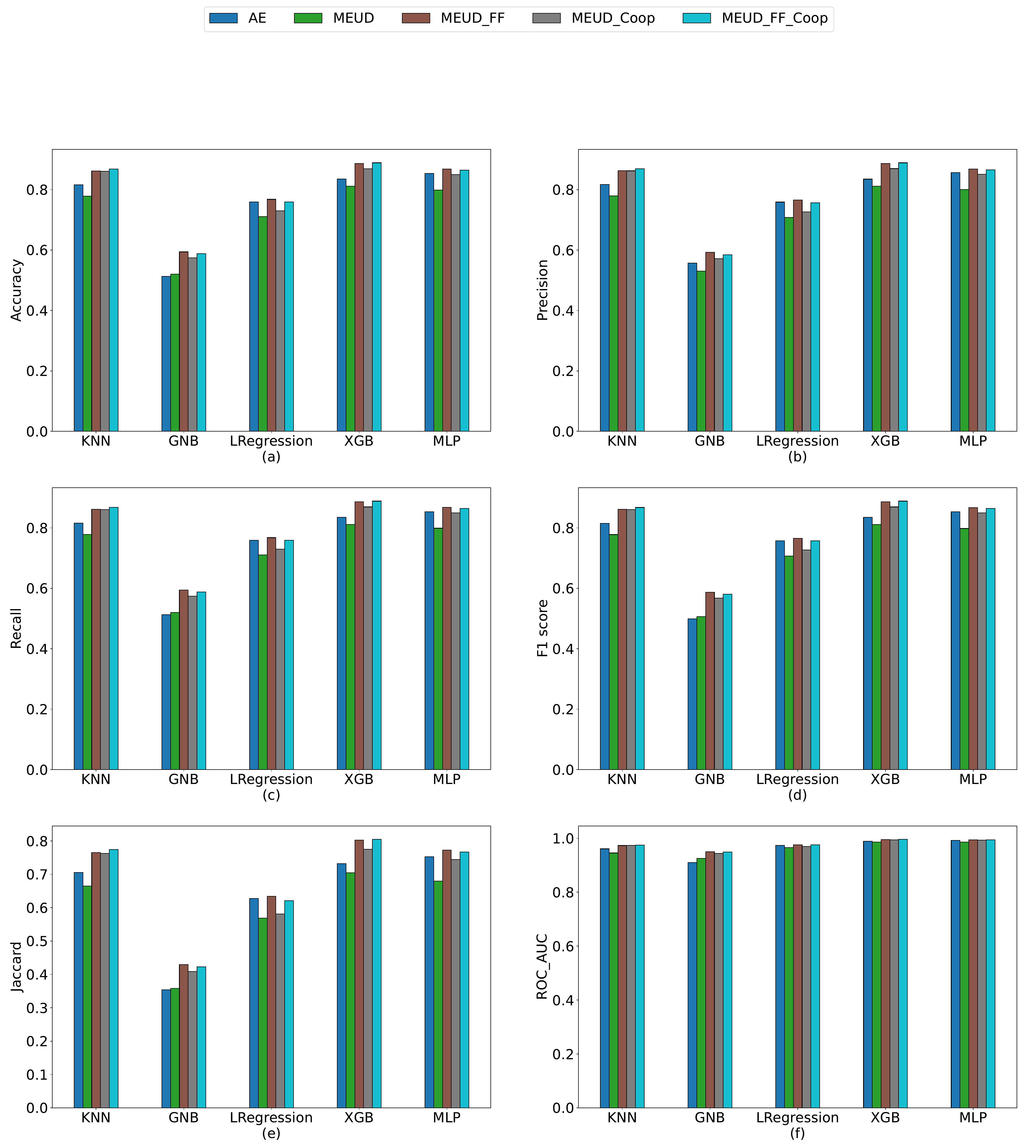}}
\caption{Mean performance scores of the classification algorithms on the dimensionally reduced dataset instances of the EMNIST dataset by MEUD-FF-Coop and four other deep learning techniques.}
\label{classify_average_EMNIST}
\end{figure}

\subsection{Results and  Analysis}
\label{analysis_meud}
The performance of MEUD-FF-Coop has been presented and justified in two parts. First, the ability to preserve the local structure of data in the low-dimensional representation has been compared in terms of the trustworthiness metric. Second, the effectiveness of the dimensionally reduced dataset is explored for downstream analyses, like classification.

\subsubsection{Quantifying the quality of low dimensional embedding: Local structure preservation}
The trustworthiness score has been computed to compare the ability of local structure preservation in the dimensionally reduced instances of the datasets by MEUD-FF-Coop against the four other models. Trustworthiness metric is used to measure the extent of local structure retention in the latent space representation of the data with respect to the original data \cite{van2009Learning, Venna2001Neighborhood, pedregosa2011scikit}. The value of trustworthiness lies in $[0, 1]$. The higher the trustworthiness score, the better is the low-rank representation, indicating the dimension reduction technique is better.

As mentioned in Section~\ref{exp_setup_meud}, each of the four datasets has been reduced to $20$ different $r$ dimension values by all five dimension reduction techniques of $T$. Thus, we obtain five different trustworthiness scores for each $r$ value. Figures~\ref{trustworthiness_MNIST}-\ref{trustworthiness_EMNIST} represent the resulting scores of the same. It can be clearly observed that the trustworthiness scores generated by MEUD-FF-Coop are better than the others in most of the cases and are clustered in a small neighbourhood. They are not scattered, and thus, justify the efficacy of MEUD-FF-Coop in preserving local structure in the low dimensional embedding. As the value of $r$ ranges from $25$ to $500$, the scalability of MEUD-FF-Coop has also been established.

Figure~\ref{trustworthiness} is a spider/star plot that depicts the average of the $20$ trustworthiness scores corresponding to twenty $r$ values. The four datasets correspond to four axes. The mean trustworthiness score of a dimension reduction technique for a particular dataset is a point on that axis. Therefore, a dimension reduction technique has four points on four axes corresponding to four datasets which can be considered as the four vertices of a polygon.  Thus, in Figure~\ref{trustworthiness}, there are five polygons for five dimension reduction techniques. The area covered by a polygon justifies the efficacy of a dimension reduction method over all the datasets together. A greater area denotes a better performance by the algorithm. The area bounded by the polygon of MEUD-FF-Coop is shown in a shaded colour in Figure~\ref{trustworthiness}, and it can be observed that it is the highest among all the five polygons. Hence, it leads to the conclusion that the quality of low dimensional embedding produced by MEUD-FF-Coop is superior in preserving the granular property of data to that produced by the other dimension reduction methods considered here.

\subsubsection{Downstream analyses: Classification}
The effectiveness of dimensionality reduction has been assessed by performing classification on the low dimensional embedding generated by MEUD-FF-Coop and the other four dimensionality reduction techniques. Different types of classification algorithms have been considered to justify a better performance of MEUD-FF-Coop over the other feature reduction techniques used here. For example, KNN is a non-parametric algorithm, the GNB algorithm belongs to the group of probabilistic classifiers, whereas MLP is a type of feed-forward artificial neural network, LRegression is a regression based classifier, and XGB is a tree based classifier. To quantify the performance of the classification algorithms, various classification metrics have been considered. This part of the experimentation aims to determine the superiority of dimension reduction by MEUD-FF-Coop using different types of classification algorithms. As mentioned earlier in Section~\ref{exp_setup_meud}, each of the four datasets has been reduced to $20$ different $r$ dimension values by all five dimension reduction techniques of $T$. Thus, for each dataset, classifier and classification performance metric, we obtain five different scores for each $r$ value corresponding to five competing models. The average of these $20$ scores has been further depicted in the plots (Figures~\ref{classify_average_MNIST}-\ref{classify_average_EMNIST}).

The classification performance of the proposed MEUD-FF-Coop along with the other four dimensionality reduction techniques are depicted in Figures~\ref{classify_average_MNIST}-\ref{classify_average_EMNIST}. Each of the five dimensionality reduction techniques undergoes classification through five classifiers and is evaluated through six classification metrics. Hence, there is a total of $30$ comparative performances of the dimensionality reduction techniques for each dataset. As observed from the figures, MEUD-FF-Coop has beaten all other dimensionality reduction techniques in all the $30$ cases. The MEUD-FF-Coop has shown similar performances for both FMNIST and CIFAR-10 datasets except once in both datasets. The MEUD-FF has been observed to have beaten the other techniques when the dimensionally reduced samples are classified through the GNB classifier. In terms of precision, the performance of the standard deep autoencoder is better than the remaining ones, when the classification method is GNB and applied to the reduced dataset instances of CIFAR-10. In the case of the EMNIST dataset, the comparative performance of the MEUD-FF-Coop is superior to the others in $13$ out of $30$ cases. MEUD-FF-Coop has performed the second best among all five dimension reduction techniques considered here when the GNB classifier has been used for all six classification performance metrics. In all of these cases, MEUD-FF has registered the best performance. A similar trend has also been observed when the classifier is MLP for all classification performance metrics other than ROC\_AUC. In the case of ROC\_AUC, MEUD-FF-Coop has the best performance for the MLP classifier. MEUD-FF-Coop has recorded the third best performance in contrast to the other four dimension reduction algorithms for the LRegression classifier for all the performance measures except ROC\_AUC. MEUD-FF and the standard deep AE models are the best and the second best dimension reduction algorithms in such cases. The performance of MEUD-FF-Coop is better than all the others in the case of the LRegression classifier when the performance is measured using ROC\_AUC.

According to the description above, it can be observed that the accuracy and recall score of the reduced dataset using MEUD-FF-Coop has surpassed the others in all the situations for MNIST, FMNIST and CIFAR-10 datasets, and five types of classifiers. The performance of MEUD-FF-Coop is either the best or very close to the leading performance for the EMNIST dataset. The supremacy of MEUD-FF-Coop is established for all occurrences of MNIST and FMNIST datasets for precision as the classification metric. The performance of MEUD-FF-Coop in CIFAR-10 and EMNIST datasets is mostly satisfactory except for a few cases, where MEUD-FF-Coop has shown comparable performance with the top performers. The MEUD-FF-Coop has been observed to have outperformed the others in terms of F1 score in all the circumstances for MNIST and CIFAR-10 datasets. In the case of FMNIST dataset, MEUD-FF-Coop has performed better in all the cases except for GNB classifier. Whereas, for EMNIST dataset, MEUD-FF-Coop has a competitive performance with others. Thus, the combination of Accuracy, Precision, Recall and F1 score justifies the superiority of MEUD-FF-Coop. MEUD-FF-Coop has outperformed the others in all the cases in terms of Jaccard index and ROC\_AUC metric for MNIST, FMNIST and CIFAR-10 datasets. The performance of MEUD-FF-Coop is mostly better than the others for EMNIST dataset. The use of different classification performance indices rationalizes the goodness of low-dimensional approximation by different dimension reduction methodologies considered here. Thus, the preceding discussion clearly demonstrates that MEUD-FF-Coop consistently produces superior low-rank representations compared to other dimension reduction techniques across various classifiers and performance metrics.

\subsection{Convergence Analysis}
\label{convergence_meud}
An important characteristic of a deep learning model is its rate of convergence, which refers to how quickly the loss curve flattens. When comparing the loss curves of different deep learning models on a specific dataset, the model that converges to the lowest loss value is generally considered superior. Additionally, the smoothness of the loss curve is a significant indicator of the learning quality of the model. A smoother curve typically reflects better model learning.

\begin{figure}[t]
\centerline{
\includegraphics[width=.66\linewidth, scale=1.0]{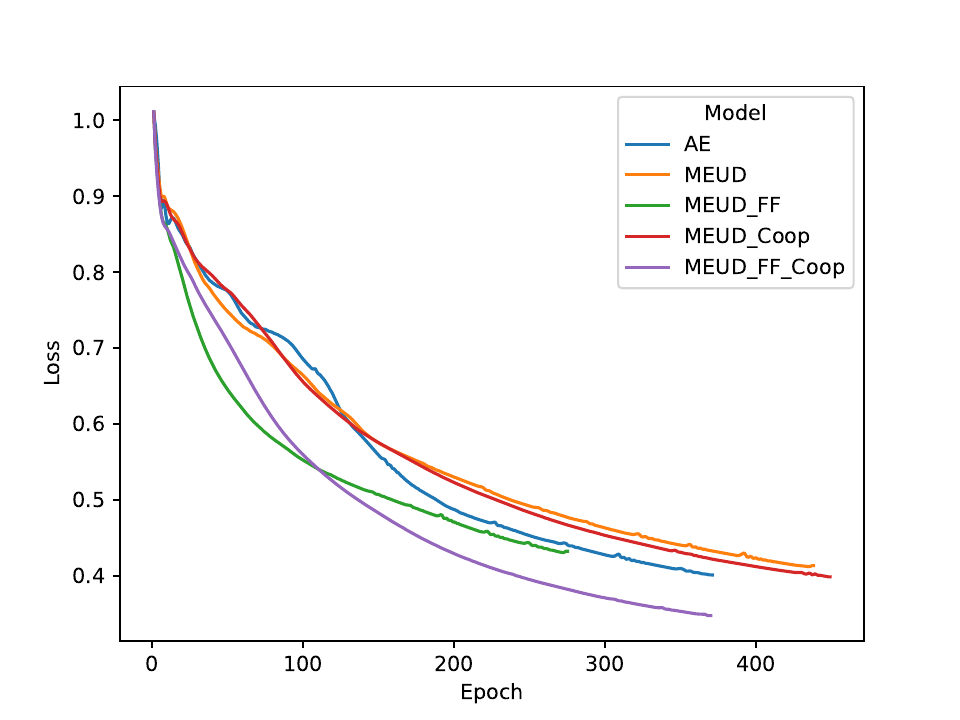}}
\caption{Loss vs. Epoch plots for various deep neural network models on the MNIST dataset.}
\label{lossCurve_250_MNIST}
\end{figure}

\begin{figure}[t]
\centerline{
\includegraphics[width=.66\linewidth, scale=1.0]{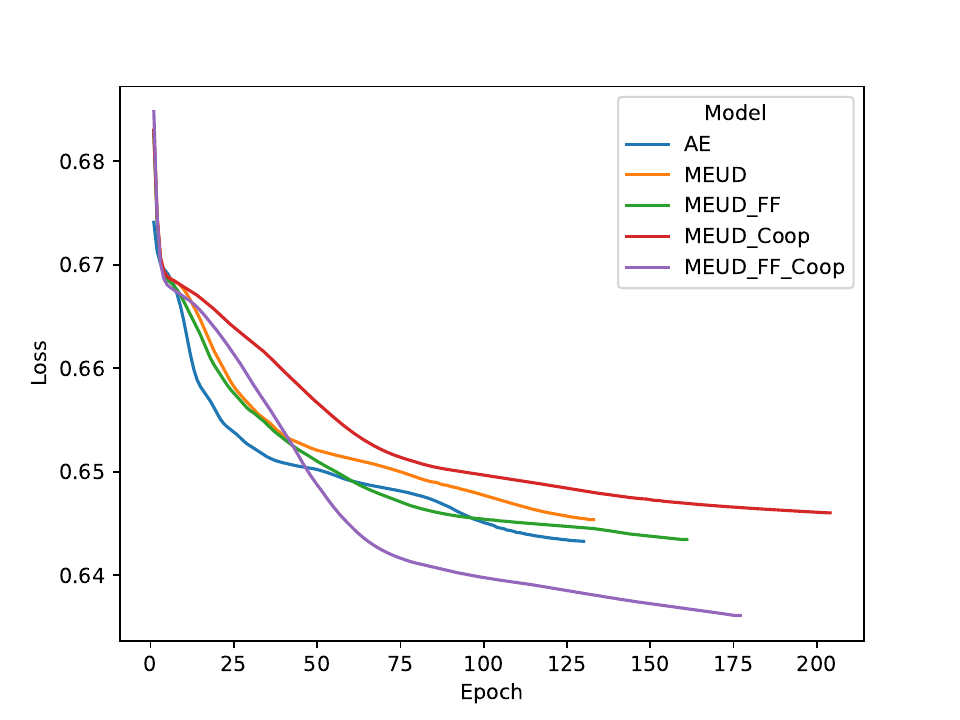}}
\caption{Loss vs. Epoch plots for various deep neural network models on the FMNIST dataset.}
\label{lossCurve_50_FMNIST}
\end{figure}

\begin{figure}[t]
\centerline{
\includegraphics[width=.66\linewidth, scale=1.0]{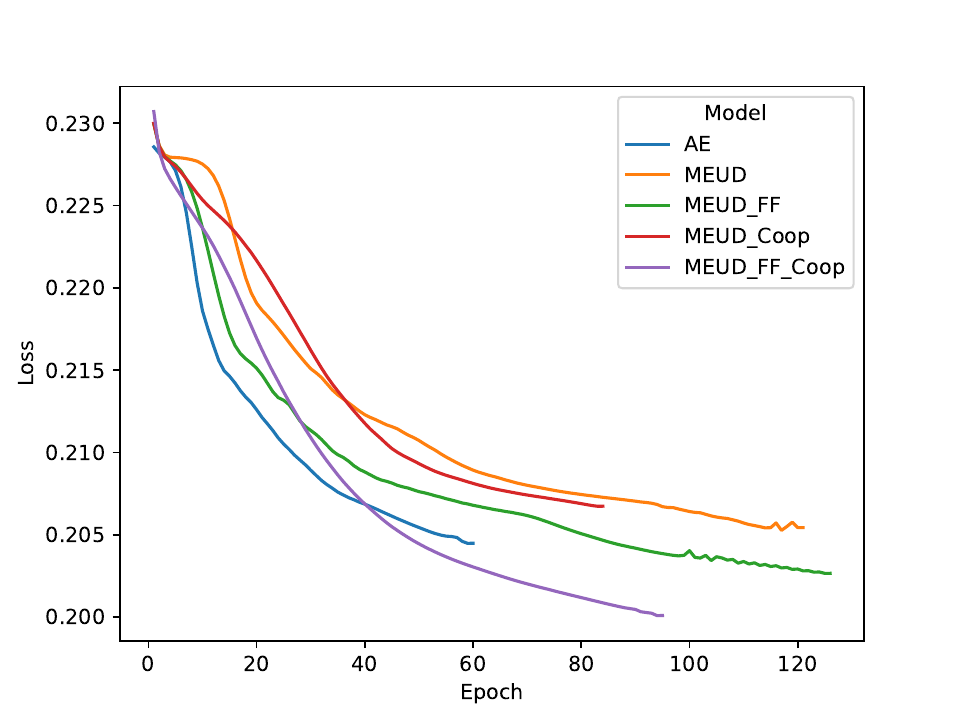}}
\caption{Loss vs. Epoch plots for various deep neural network models on the CIFAR-10 dataset.}
\label{lossCurve_50_CIFAR10}
\end{figure}

\begin{figure}[t]
\centerline{
\includegraphics[width=.66\linewidth, scale=1.0]{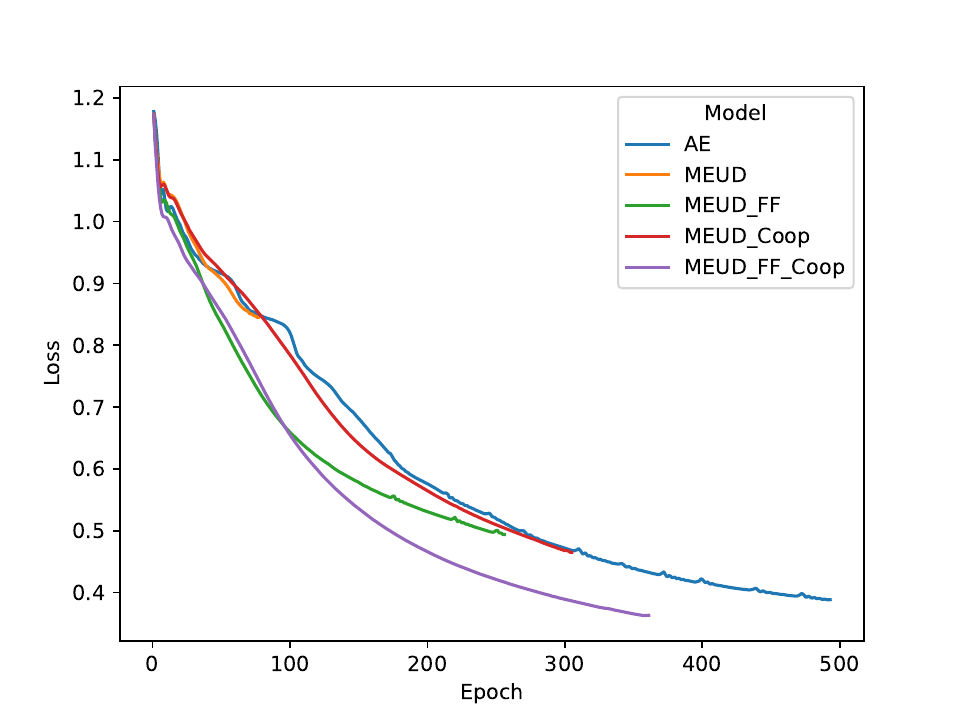}}
\caption{Loss vs. Epoch plots for various deep neural network models on the EMNIST dataset.}
\label{lossCurve_200_EMNIST}
\end{figure}

Figures~\ref{lossCurve_250_MNIST}-\ref{lossCurve_200_EMNIST} depict the Loss vs. Epoch plots for all four datasets considered here for a particular $r$ value. Each figure displays the loss curves of all five deep learning models included in this comparative analysis. For a given dataset, all five dimensionality reduction models have reduced the input to the same target dimension, $r$. The value of the reduced dimension for MNIST dataset is $250$, for both FMNIST and CIFAR-10, it is $50$ and for EMNIST the same value is $200$.  Across all the four datasets, MEUD-FF-Coop model has consistently achieved the lowest loss value. Additionally, MEUD-FF-Coop has demonstrated the fastest convergence to this minimum loss for each dataset. The loss curves generated by MEUD-FF-Coop also show a well-defined elbow shape, indicating superior convergence behaviour compared to the other models. In terms of loss curve smoothness, MEUD-FF-Coop has again outperformed the other five deep learning models, exhibiting the smoothest curve. This establishes that the MEUD-FF-Coop model has demonstrated superior convergence across multiple aspects. This not only establishes the efficacy of the MEUD-FF-Coop neural network model but also justifies the FCB learning algorithm developed here mimicking the human nature of learning a new topic.

\section{Conclusion}
\label{conclusion_meud}
Learning algorithms have been developed to emulate human-like learning through machines. The backpropagation method of learning has played a major role in the development of neural networks ever since its inception. Newer techniques, like the Forward Forward learning, have also tried to develop a more biologically plausible learning mechanism. However, such techniques are not completely able to mimic the human nature of learning. In this paper, we have developed a human-centric learning mechanism that combines forward-forward learning, backpropagation, and cooperation. The cooperation technique establishes the concept of discussion among peers to enhance one's knowledge. Thus, we have come up with Forward-Cooperation-Backward (FCB) learning which tries to mimic the human nature of learning similar to a student trying to grasp a new concept. FCB learning strategy has gradually been established through the MEUD, MEUD-FF, MEUD-Coop and MEUD-FF-Coop frameworks. The complete realization of FCB learning algorithm has been accomplished using the novel MEUD-FF-Coop neural network model.

To establish the superiority of MEUD-FF-Coop over others, the dimensionally reduced data obtained from the network have been tested using two ways, viz., the trustworthiness score, i.e., the ability to preserve the granular relationship of data and the classification performance on the low dimensional embedding produced by the algorithms. The trustworthiness scores and its depiction through the spider/star plot justify the efficiency of MEUD-FF-Coop over the others. It is seen that the trustworthiness scores generated by MEUD-FF-Coop are clustered together and are mostly better than others. The area coverage in the spider plot by the polygon relating to MEUD-FF-Coop is the highest, in turn justifying the superiority of the FCB learning technique. Different classifiers and classification performance metrics have been used to justify the efficacy of dimension reduction as downstream analyses establishing FCB learning procedure. For three out of four datasets, the performance of MEUD-FF-Coop has been found to be better among the others in terms of classification, and displays a competitive performance for the EMNIST dataset. The experimental convergence analyses show that MEUD-FF-Coop converges the fastest among the others and is able to reduce the loss to the lowest. Additionally, the smoothness of the convergence curve in the case of MEUD-FF-Coop is the highest. 

Thus, the FCB way of learning establishes its superiority along with the novel MEUD-FF-Coop neural network framework in terms of the preservation of local structure in the transformed dataset and downstream analyses using classification. The convergence analysis of the models also supports the efficacy of the developed FCB learning technique.


\end{document}